# Expectation-Propagation for the Generative Aspect Model

**Thomas Minka**
Department of Statistics
Carnegie Mellon University
Pittsburgh, PA 15213 USA
minka@stat.cmu.edu

**John Lafferty**
School of Computer Science
Carnegie Mellon University
Pittsburgh, PA 15213 USA
lafferty@cs.cmu.edu

**Abstract**

The generative aspect model is an extension of the multinomial model for text that allows word probabilities to vary stochastically across documents. Previous results with aspect models have been promising, but hindered by the computational difficulty of carrying out inference and learning. This paper demonstrates that the simple variational methods of Blei et al. (2001) can lead to inaccurate inferences and biased learning for the generative aspect model. We develop an alternative approach that leads to higher accuracy at comparable cost. An extension of Expectation-Propagation is used for inference and then embedded in an EM algorithm for learning. Experimental results are presented for both synthetic and real data sets.

## 1 Introduction

Approximate inference techniques, such as variational methods, are increasingly being used to tackle advanced data models. When learning and inference are intractable, approximation can make the difference between a useful model and an impractical model. However, if applied indiscriminately, approximation can change the qualitative behavior of a model, leading to unpredictable and undesirable results.

The generative aspect model introduced by Blei et al. (2001) is a promising model for discrete data, and provides an interesting example of the need for good approximation strategies. When applied to text, the model explicitly accounts for the intuition that a document may have several subtopics or "aspects," making it an attractive tool for several applications in text processing and information retrieval. As an example, imagine that rather than simply returning a list of documents that are "relevant" to a given topic, a search engine could automatically determine the different aspects of the topic that are treated by each document in the list, and reorder the documents to efficiently cover these different aspects. The TREC interactive track (Over, 2001) has been set up to help investigate precisely this task, but from the point of view of users interacting with the search engine. As an example of the judgements made in this task, for the topic "electric automobiles" (number 247i), the human assessors identified eleven aspects among the documents that were judged to be relevant, having descriptions such as "government funding of electric car development programs," "industrial development of hybrid electric cars," and "increased use of aluminum bodies."

This paper examines computation in the generative aspect model, proposing new algorithms for approximate inference and learning that are based on the Expectation-Propagation framework of Minka (2001b). Hofmann's original aspect model involved a large number of parameters and heuristic procedures to avoid overfitting (Hofmann, 1999). Blei et al. (2001) introduced a modified model with a proper generative semantics and used variational methods to carry out inference and learning. It is found that the variational methods can lead to inaccurate inferences and biased learning, while Expectation-Propagation gives results that are more true to the model. Besides providing a practical new algorithm for a useful model, we hope that this result will shed light on the question of which approximations are appropriate for which problems.

The following section presents the generative aspect model, briefly discussing some of the properties that make it attractive for modeling documents, and stating the inference and learning problems to be addressed. After a brief overview of Expectation-Propagation in Section 3, a new algorithm for approximate inference in the generative aspect model is presented in Section 4. Separate from Expectation-Propagation, a new algorithm for approximate learning in the generative aspect model is presented in Section 5. Brief descriptions of the corresponding procedures with variational methods are included for completeness. Section 6 describes experiments on synthetic and real data. Section 6.1 presents a synthetic data experiment us-ing low dimensional multinomials, which clearly demonstrates how variational methods can result in inaccurate inferences compared to Expectation-Propagation. In Sections 6.2 and 6.3 the methods are then compared using document collections taken from TREC data, where it is seen that Expectation-Propagation attains lower test set perplexity. Section 7 summarizes the results of the paper.

## 2 The Generative Aspect Model

A simple generative model for documents is the multinomial model, which assumes that words are drawn one at a time and independently from a fixed word distribution $p(w)$. The probability of a document $d$ having word counts $n_w$ is thus

$$p(d \mid \boldsymbol{p}) = \prod_{w=1}^{W} p(w)^{n_w} \quad (1)$$

This family is very restrictive, in that a document of length $n$ is expected to have $np(w)$ occurrences of word $w$, with little variation away from this number. Even within a homogeneous set of documents, such as machine learning papers, there is typically far more variation in the word counts. One way to accommodate this is to allow the word probabilities $\boldsymbol{p}$ to vary across documents, leading to a hierarchical multinomial model.

This requires us to specify a distribution on $\boldsymbol{p}$ itself, considered as a vector of numbers which sum to one. One natural choice is the Dirichlet distribution, which is conjugate to the multinomial. Unfortunately, while the Dirichlet can capture variation in the $p(w)$'s, it cannot capture *co-variation*, the tendency for some probabilities to move up and down together. At the other extreme, we can sample $\boldsymbol{p}$ from a finite set, corresponding to a finite mixture of multinomials. This model can capture co-variation, but at great expense, since a new mixture component is needed for every distinct choice of word probabilities.

In the generative aspect model, it is assumed that there are $A$ underlying aspects, each represented as a multinomial distribution over the words in the vocabulary. A document is generated by the following process. First, $\lambda$ is sampled from a Dirichlet distribution $\mathcal{D}(\lambda \mid \alpha)$, so that $\sum_a \lambda_a = 1$. This determines mixing weights for the aspects, yielding a word probability vector:

$$p(w \mid \lambda) = \sum_a p(w \mid a) \lambda_a \quad (2)$$

The document is then sampled from a multinomial distribution with these probabilities. Instead of a finite mixture, this distribution might be called a *simplicial mixture*, since the word probability vector ranges over a simplex with corners $p(w \mid a = 1), ..., p(w \mid a = A)$. The probability of a document is

$$p(d \mid \theta) = \int_\Delta \mathcal{D}(\lambda \mid \alpha) \prod_{w=1}^{W} \left( \sum_a \lambda_a p(w \mid a) \right)^{n_w} d\lambda \quad (3)$$

where the parameters $\theta$ are the Dirichlet parameters $\alpha_a$ and the multinomial models $p(\cdot \mid a)$; $\Delta$ denotes the $(A-1)$-dimensional simplex, the sample space of the Dirichlet $\mathcal{D}(\cdot \mid \alpha)$. Because $\lambda$ is sampled for each document, different documents can exhibit the aspects in different proportions. However, the integral in (3) does not simplify and must be approximated, which is the main complication in using this model.

The two basic computational tasks for this model are:

**Inference:** Evaluate the probability of a document; *i.e.*, the integral in (3).

**Learning:** For a set of training documents, find the parameter values $\theta = (p(\cdot \mid a), \alpha)$ which maximize the likelihood; *i.e.*, maximize the value of the integral in (3).

## 3 Expectation-Propagation

Expectation-Propagation is an algorithm for approximating integrals over functions that factor into simple terms. The general form for such integrals in our setting is

$$\int_\Delta p(\lambda) \prod_{w=1}^{W} t_w(\lambda)^{n_w} d\lambda \quad (4)$$

In previous work each count $n_w$ was assumed to be 1 (Minka, 2001b). Here we present a slight generalization to allow real-valued powers on the terms. Expectation-Propagation approximates each term $t_w(\lambda)$ by a simpler term $\tilde{t}_w(\lambda)$, giving a simpler integral

$$\int_\Delta q(\lambda) \, d\lambda, \qquad q(\lambda) = p(\lambda) \prod_{w=1}^{W} \tilde{t}_w(\lambda)^{n_w} \quad (5)$$

whose value is used to estimate the original.

The algorithm proceeds by iteratively applying "deletion/inclusion" steps. One of the approximate terms is deleted from $q(\lambda)$, giving the partial function $q^{\setminus w}(\lambda) = q(\lambda)/\tilde{t}_w(\lambda)$. Then a new approximation for $t_w(\lambda)$ is computed so that $t_w(\lambda) \, q^{\setminus w}(\lambda)$ is similar to $\tilde{t}_w(\lambda) \, q^{\setminus w}(\lambda)$, in the sense of having the same integral and the same set of specified moments. The moments used in this paper are the mean and variance. The partial function $q^{\setminus w}(\lambda)$ thus acts as context for the approximation. Unlike variational bounds, this approximation is global, not local, and consequently the estimate of the integral is more accurate.

A fixed point of this algorithm always exists, but we may not always reach one. The approximation may oscillate or enter a region where the integral is undefined. We utilize two techniques to prevent this. First, the updates are "damped" so that $\tilde{t}_w(\lambda)$ cannot oscillate. Second, if a deletion-inclusion step leads to an undefined integral, the step is undone and the algorithm continues with the next term.

## 4 Inference

This section describes two algorithms for approximating the integral in (3): variational inference used by Blei et al. (2001) and Expectation-Propagation.

### 4.1 Variational inference

To approximate the integral of a function, variational inference lower bounds the function and then integrates the lower bound. A simple lower bound for (3) comes from Jensen's inequality. The bound is parameterized by a vector $q(a \mid w)$:

$$\sum_a \lambda_a p(w \mid a) \geq \prod_a \left(\frac{\lambda_a p(w \mid a)}{q(a \mid w)}\right)^{q(a \mid w)} \quad (6)$$

$$\sum_a q(a \mid w) = 1 \quad (7)$$

The vector $q(a \mid w)$ can be interpreted as a soft assignment or "responsibility" of word $w$ to the aspects. Given bound parameters $q(a \mid w)$ for all $a$ and $w$, the integral is now analytic:

$$p(d \mid \theta) \geq$$
$$\prod_{wa} \frac{p(w \mid a)^{n_w q(a \mid w)}}{q(a \mid w)} \int_\Delta \mathcal{D}(\lambda \mid \alpha) \prod_a \lambda_a^{\sum_w n_w q(a \mid w)} d\lambda \quad (8)$$

$$\int_\Delta \mathcal{D}(\lambda \mid \alpha) \prod_a \lambda_a^{\sum_w n_w q(a \mid w)} d\lambda =$$
$$\frac{\prod_a \Gamma(\alpha_a + \sum_w n_w q(a \mid w))}{\Gamma(\sum_a \alpha_a + n)} \frac{\Gamma(\sum_a \alpha_a)}{\prod_a \Gamma(\alpha_a)} \quad (9)$$

The best bound parameters are found by maximizing the value of the bound. A convenient way to do this is with EM. The "parameter" in the algorithm is $q(a \mid w)$ and the "hidden variable" is $\lambda$:

**E-step:** $\quad \gamma_a = \alpha_a + \sum_w n_w q(a \mid w) \quad (10)$

**M-step:** $\quad q(a \mid w) \propto p(w \mid a) \exp(\Psi(\gamma_a)) \quad (11)$

where $\Psi$ is the digamma function. Note that this is the same as the variational algorithm for mixture weights given by Minka (2000). The variables $\gamma_a$ used in this algorithm can be interpreted as defining an approximate Dirichlet posterior on $\lambda$: $\mathcal{D}(\lambda \mid \gamma)$.

### 4.2 Expectation-Propagation

As mentioned above, the aspect model integral is the same as marginalizing the weights of a mixture model whose components are known. This problem was studied in depth by Minka (2001b) and it was found that Expectation-Propagation (EP) provides the best results, compared to Laplace's method using a softmax transformation, variational inference, and two different Monte Carlo algorithms.

EP gives an integral estimate as well as an approximate posterior for the mixture weights. For the generative aspect model, the approximate posterior will be Dirichlet, and the integrand will be factored into terms of the form

$$t_w(\lambda) = \sum_a p(w \mid a) \lambda_a \quad (12)$$

so that the integral we want to solve is

$$\int_\Delta \mathcal{D}(\lambda \mid \alpha) \prod_{w=1}^W t_w(\lambda)^{n_w} d\lambda \quad (13)$$

To apply EP, the term approximations are taken to have a product form,

$$\tilde{t}_w(\lambda) = s_w \prod_a \lambda_a^{\beta_{wa}} \quad (14)$$

which resembles a Dirichlet with parameters $\beta_{wa}$. Thus, the approximate posterior is given by

$$q(\lambda) \propto \mathcal{D}(\lambda \mid \alpha) \prod_w \tilde{t}_w(\lambda)^{n_w} = \mathcal{D}(\lambda \mid \gamma) \quad (15)$$

$$\text{where } \gamma_a = \alpha_a + \sum_w n_w \beta_{wa} \quad (16)$$

To begin EP, the parameters are initialized by setting $\beta_{wa} = 0$, $s_w = 1$. Because the $\tilde{t}_w$ are initialized to 1, this starts out as the prior: $\gamma_a = \alpha_a$.

EP then iteratively passes through the words in the document, performing the following steps until all $(\beta_w, s_w)$ converge:

loop $w = 1, ..., W$:

(a) *Deletion.* Remove $\tilde{t}_w$ from the posterior to get an "old" posterior:

$$\gamma_a^{\setminus w} = \gamma_a - \beta_{wa} \quad (17)$$

If any $\gamma^{\setminus w} < 0$, skip this word for this iteration of EP.

(b) *Moment matching.* Compute $\gamma'$ from $\gamma^{\setminus w}$ by matching the mean and variance of the corresponding Dirichlet distributions; see equations (23–25).

(c) *Update.* Reestimate $\tilde{t}_w$ using stepsize $\mu$:

$$\beta_{wa} = \mu(\gamma'_a - \gamma_a^{\setminus w}) + (1 - \mu)\beta_{wa}^{old} \quad (18)$$

$$Z_w(\gamma^{\setminus w}) = \frac{\sum_a p(w \mid a)\gamma_a^{\setminus w}}{\sum_a \gamma_a^{\setminus w}} \quad (19)$$

$$s_w = Z_w(\gamma^{\setminus w}) \frac{\Gamma(\sum_a \gamma'_a)}{\prod_a \Gamma(\gamma'_a)} \frac{\prod_a \Gamma(\gamma_a^{\setminus w})}{\Gamma(\sum_a \gamma_a^{\setminus w})} \quad (20)$$

(d) *Inclusion*. Incorporate $\bar{t}_w$ back into $q(\boldsymbol{\lambda})$ by scaling the change in $\beta$:

$$\gamma_a = \gamma_a^{old} + n_w(\beta_{wa} - \beta_{wa}^{old}) \quad (21)$$

This preserves the invariant (16). If any $\gamma_a < 0$, undo all changes and skip this word.

In our experience, words are skipped only on the first few iterations, before EP has settled into a decent approximation. It can be shown that the safest stepsize for (c) is $\mu = 1/n_w$, which makes $\gamma = \gamma'$. This is the value used in the experiments, though for faster convergence a larger $\mu$ is often acceptable.

After convergence, the approximate posterior gives the following estimate for the likelihood of the document, thus approximating the integral (3):

$$Z(d) = \frac{\prod_a \Gamma(\gamma_a)}{\Gamma(\sum_a \gamma_a)} \frac{\Gamma(\sum_a \alpha_a)}{\prod_a \Gamma(\alpha_a)} \prod_{w=1}^{W} s_w^{n_w} \quad (22)$$

A calculation shows that the mean and variance of the Dirichlets are matched in step (b) by using the following update to $\gamma_a$ (Cowell et al., 1996):

$$m_a = \frac{1}{Z_w(\gamma^{\setminus w})} \frac{\gamma_a^{\setminus w}}{\sum_a \gamma_a^{\setminus w}} \frac{p(w \mid a) + \sum_a p(w \mid a) \gamma_a^{\setminus w}}{1 + \sum_a \gamma_a^{\setminus w}} \quad (23)$$

$$m_a^{(2)} = \frac{1}{Z_w(\gamma^{\setminus w})} \frac{\gamma_a^{\setminus w}}{\sum_a \gamma_a^{\setminus w}} \frac{\gamma_a^{\setminus w} + 1}{\sum_a \gamma_a^{\setminus w} + 1} \times$$
$$\frac{2 p(w \mid a) + \sum_a p(w \mid a) \gamma_a^{\setminus w}}{2 + \sum_a \gamma_a^{\setminus w}} \quad (24)$$

$$\gamma_a' = \left( \frac{\sum_a (m_a - m_a^{(2)})}{\sum_a (m_a^{(2)} - m_a^2)} \right) m_a \quad (25)$$

## 5　Learning

Given a set of documents $\mathcal{C} = \{d_i, i = 1, ..., n\}$, with word counts denoted $n_{iw}$, the learning problem is to maximize the likelihood as a function of the parameters $\theta = (p(\cdot \mid a), \alpha)$; the likelihood is given by

$$p(\mathcal{C} \mid \theta) = \prod_i \int_\Delta \mathcal{D}(\boldsymbol{\lambda} \mid \alpha) \prod_{w=1}^{W} \left( \sum_a \lambda_a p(w \mid a) \right)^{n_{iw}} d\boldsymbol{\lambda} \quad (26)$$

Notice that each document has its own integral over $\boldsymbol{\lambda}$. It is tempting to use EM for this problem, where we regard $\boldsymbol{\lambda}$ as a hidden variable for each document. However, the E-step requires expectations over the posterior for $\boldsymbol{\lambda}$, $p(\boldsymbol{\lambda} \mid d_i, \theta)$, which is an intractable distribution. This section describes two alternative approaches: (1) maximizing the likelihood estimates from the previous section and (2) a new approach based on approximative EM. The decision between these two approaches is separate from the decision of using variational inference versus Expectation-Propagation.

### 5.1　Maximizing the estimate

Given that we can estimate the likelihood function for each document, it seems natural to try to maximize the value of the estimate. This is the approach taken by Blei et al. (2001). For the variational bound (8), the maximum with respect to the parameters is obtained at

$$p(w \mid a)^{new} \propto \sum_i n_{iw} q_i(a \mid w) \quad (27)$$

$$\alpha^{new} = \arg\max_\alpha \prod_i \frac{\prod_a \Gamma(\alpha_a + \sum_w n_{iw} q_i(a \mid w))}{\Gamma(\sum_a \alpha_a + \sum_w n_{iw})} \frac{\Gamma(\sum_a \alpha_a)}{\prod_a \Gamma(\alpha_a)} \quad (28)$$

Of course, once the aspect parameters are changed, the optimal bound parameters $q(a \mid w)$ also change, so Blei et al. (2001) alternate between optimizing the bound and applying these updates. This can be understood as an EM algorithm where both $\boldsymbol{\lambda}$ and the 'aspect assignments' are hidden variables. The aspect parameters at convergence will result in the largest possible variational estimate of the likelihood. The same approach could be taken with EP, where we find the parameters that result in the largest possible EP estimate of the likelihood. However, this does not seem to be as simple as in the variational approach. It also seems misguided, because an approximation which is close to the true likelihood in an average sense need not have its maximum close to the true maximum.

### 5.2　Approximative EM

The second approach is to use an approximative EM algorithm, sometimes called "variational EM," where we use expectations over an approximate posterior for $\boldsymbol{\lambda}$, call it $q_i(\boldsymbol{\lambda})$. The inference algorithms in the previous section conveniently give such an approximate posterior. The E-step will compute $q_i(\boldsymbol{\lambda})$ for each document, and the M-step will maximize the following lower bound to the log-likelihood:

$$\log p(\mathcal{C} \mid \theta) \geq$$
$$\sum_i \int_\Delta q_i(\boldsymbol{\lambda}) \log \mathcal{D}(\boldsymbol{\lambda} \mid \alpha) \prod_{w=1}^{W} \left( \sum_a \lambda_a p(w \mid a) \right)^{n_{iw}} d\boldsymbol{\lambda}$$
$$- \sum_i \int_\Delta q_i(\boldsymbol{\lambda}) \log q_i(\boldsymbol{\lambda}) \, d\boldsymbol{\lambda} \quad (29)$$
$$= \int_\Delta \left( \sum_i q_i(\boldsymbol{\lambda}) \right) \log \mathcal{D}(\boldsymbol{\lambda} \mid \alpha) \, d\boldsymbol{\lambda} \; +$$

$$\sum_{iw} n_{iw} \int_{\Delta} q_i(\boldsymbol{\lambda}) \log \left( \sum_a \lambda_a p(w \,|\, a) \right) d\boldsymbol{\lambda} + \text{const.}$$

This decouples into separate maximization problems for $\alpha$ and $p(w \,|\, a)$. Given that $q_i(\boldsymbol{\lambda})$ is Dirichlet with parameters $\gamma_{ia}$, the optimization problem for $\alpha$ is to maximize

$$\begin{aligned}
& n \left( \log \Gamma(\sum_a \alpha_a) - \sum_a \log \Gamma(\alpha_a) \right) \\
& + \sum_i \sum_a (\alpha_a - 1) E_q[\log \lambda_{ia}] \\
= & \, n \left( \log \Gamma(\sum_a \alpha_a) - \sum_a \log \Gamma(\alpha_a) \right) \\
& + \sum_{ia} (\alpha_a - 1) \left( \Psi(\gamma_{ia}) - \Psi(\sum_a \gamma_{ia}) \right) \quad (30)
\end{aligned}$$

which is the standard Dirichlet maximum-likelihood problem (Minka, 2001a).

By zeroing the derivative with respect to $p(w \,|\, a)$, we obtain the M-step

$$p(w \,|\, a)^{new} \propto \sum_i n_{iw} \int_{\Delta} q_i(\boldsymbol{\lambda}) \frac{\lambda_a p(w \,|\, a)}{\sum_a \lambda_a p(w \,|\, a)} d\boldsymbol{\lambda} \quad (31)$$

This requires approximating another integral over $\boldsymbol{\lambda}$. Update (27) is equivalent to assuming that $\lambda_a$ is constant, at the value $\exp(\Psi(\gamma_a))$ (from (11)). A more accurate approximation can be obtained by Taylor expansion, as described in the appendix. The resulting update is

$$m_{iab} = \frac{\gamma_{ib} + \delta(a - b)}{\sum_b \gamma_{ib} + 1} \quad (32)$$

$$S_{ia} = \frac{\sum_b p(w \,|\, b)^2 m_{iab}}{(\sum_b p(w \,|\, b) m_{iab})^2} - 1 \quad (33)$$

$$p(w \,|\, a)^{new} \propto$$
$$\sum_i n_{iw} p(w \,|\, a) \frac{\gamma_{ia}}{\sum_b \gamma_{ib}} \frac{1}{\sum_b p(w \,|\, b) m_{iab}} \times$$
$$\left( 1 + \frac{S_{ia}}{\sum_b \gamma_{ib} + 2} \right) \quad (34)$$

This update can be used with the $\gamma$'s found by either VB or EP. When running EP with the new parameter values, the $\beta$'s can be started from their previous values, so that only a few EP iterations are required.

# 6 Experimental Results

This section presents the result of experiments carried out on synthetic and real data. The first experiments involve a "toy" data set where the aspects are multinomials over a two word vocabulary. Later experiments use documents from two TREC collections.

## 6.1 Results on Synthetic Data

This section elucidates the difference between variational inference (VB) and Expectation Propagation (EP) using simple, controlled datasets. The algorithms mainly differ in how they approximate (3), thus it is helpful to consider two extremes: (Exact) the exact value of (3) versus (Max) approximating by the maximum over $\boldsymbol{\lambda}$:

$$\int_{\Delta} \mathcal{D}(\boldsymbol{\lambda} \,|\, \boldsymbol{\alpha}) \prod_{w=1}^W \left( \sum_a \lambda_a p(w \,|\, a) \right)^{n_w} d\boldsymbol{\lambda} \approx$$
$$\mathcal{D}(\hat{\boldsymbol{\lambda}} \,|\, \boldsymbol{\alpha}) \prod_{w=1}^W \left( \sum_a \hat{\lambda}_a p(w \,|\, a) \right)^{n_w} \times \text{(constant)} \quad (35)$$

Under this approximation, the aspect parameters $p(w \,|\, a)$ only serve to restrict the domain of the word probabilities $p(w) = \sum_a \hat{\lambda}_a p(w \,|\, a)$. To maximize likelihood, we would want the domain to be as large as possible—the aspects as extreme and distinct as possible. However, when using the exact value of (3), all choices of $\boldsymbol{\lambda}$ contribute, which favors a domain that *only* includes word probabilities matching the frequencies in the documents. In experiments, we find that VB behaves like the Max approximation while EP behaves like the exact value.

Consider a simple scenario in which there are only two words in the vocabulary, $w=1$ and $w=2$. This allows each aspect to be represented by one parameter $p(w=1 \,|\, a)$, since $p(w=2 \,|\, a) = 1 - p(w=1 \,|\, a)$. Let there be two aspects, $a=1$ and $a=2$, with $\alpha_1=\alpha_2=1$, so that $\mathcal{D}(\boldsymbol{\lambda} \,|\, \boldsymbol{\alpha})$ is uniform. This means that the probability of word 1 in the document collection varies uniformly between $p(w=1 \,|\, a=1)$ and $p(w=1 \,|\, a=2)$. Learning the aspects from data amounts to estimating the endpoints of this variation.

Let $p(w=1 \,|\, a=2) = 1$ so that the only free parameter is $p(w=1 \,|\, a=1)$. Ten training documents of length 10 are generated from the model with $p(w=1 \,|\, a=1) = 0.5$. Figure 1 (top) shows the typical result. When we apply the Max approximation, each document $i$ wants to choose $p(w=1)$ (between $p(w=1 \,|\, a=1)$ and $p(w=1 \,|\, a=2)$) to match its frequency of word 1: $n_{i1}/n_i$. Any choice of $(p(w=1 \,|\, a=1), p(w=1 \,|\, a=2))$ which spans these frequencies will maximize likelihood, e.g. $p(w=1 \,|\, a=1) = 0$, $p(w=1 \,|\, a=2) = 1$. The exact likelihood, by contrast, peaks near the true value of $p(w=1 \,|\, a=1)$. As the number of training documents increases, the exact likelihood gets sharper around the true value, but the Max approximation gets *farther away* from the truth, because the observed frequencies exhibit more variance.

As shown in Figure 1 (bottom), VB behaves similarly to Max. The solid curve is the exact likelihood, generated by computing the probability of the training documents for all $p$'s on a fine grid. The dashed curve is the VB estimate of the likelihood, scaled up to make its shape visible on the plot, for the same $p$'s. The dot-dashed curve is the EP

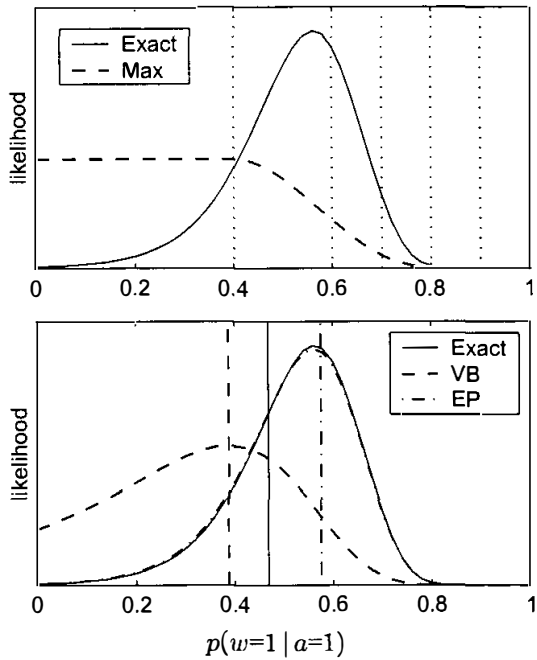

Figure 1: (Top) The exact likelihood for $p(w=1 \mid a=1)$ and its Max approximation. The observed frequencies of word 1, $n_{i1}/n_i$, are shown as vertical lines (some are identical). The Max approximation is highest when below the smallest observed frequency (0.4). (Bottom) The VB likelihood is similar to Max. The EP likelihood is nearly exact. Parameter estimates are shown as vertical lines.

estimate of the likelihood for the same $p$'s. EP clearly gives a better approximation.

The parameter estimates are indicated by vertical lines. The dashed vertical line corresponds to Blei et al's algorithm, which as expected converges to the maximum of the VB curve. The solid line is the result of VB combined with the EM update of Section 5.2; as expected, it is closer to the true maximum. The dot-dash vertical line is the result of applying EM using EP and is closest to the true maximum.

To demonstrate the difference between the algorithms in the multidimensional case, 100 documents of length 100 were generated from a simple multinomial model with five words having equal probability. A generative aspect model was fit using three aspects with $\alpha_1=\alpha_2=\alpha_3=1$. The EP solution correctly chose all aspects to be similar to the generating multinomial; all probabilities were between 0.15 and 0.24. The VB solution is quite different; it chose the extreme parameters shown in the following table (rounded to the tenths place):

|     | $w=1$ | $w=2$ | $w=3$ | $w=4$ | $w=5$ |
|-----|-------|-------|-------|-------|-------|
| $a=1$ | 0 | 0 | 0 | 0.6 | 0.4 |
| $a=2$ | 0.6 | 0.2 | 0.1 | 0 | 0.1 |
| $a=3$ | 0 | 0.4 | 0.5 | 0 | 0.1 |

Resampling the training documents gives similar results. In terms of convergence rate, learning typically converged after 150 parameter updates with EP, while over 1,000 updates were required with VB.

Interestingly, on an independent set of 1000 test documents, the perplexity of the model learned by EP is 5.0 while the VB model's is 5.1, a seemingly trivial difference. This is because the perplexity measure, as used by Blei et al., focuses on per-word prediction rather than per-document prediction. As long as there exists a mixture of the aspects which matches the word probabilities in the document, the perplexity will be low. Indeed, if the above VB aspects are evenly mixed, the correct word distribution is produced.

To show that there really is a difference between the models, a synthetic classification problem was constructed. One class had documents sampled from a uniform multinomial over five words. The other class had documents sampled from a multinomial with word probabilities [1 2 3 4 5]/15. There were 50 documents of length 50 in each class. A three-aspect model was trained on each class and test documents (of the same length) were classified according to highest class-conditional probability. The EP models committed 76/2000 errors while the VB models committed 163/2000, which is both statistically and practically significant. As above, EP learned the correct models while VB chose extreme probabilities.

### 6.2 Controlled TREC Data

In order to compare variational inference and Expectation-Propagation on more realistic data, a corpus was created by mixing together TREC documents on known topics. From the 1989 AP data on TREC disks 1 and 2, we extracted all of the documents that were judged to be relevant to one of the following six topics:

| Topic 20 | Patent Infringement Lawsuits |
| Topic 59 | Weather Related Fatalities |
| Topic 67 | Politically Motivated Civil Disturbances |
| Topic 85 | Official Corruption |
| Topic 110 | Black Resistance Against the South African Government |
| Topic 142 | Impact of Government Regulated Grain Farming on International Relations |

Synthetic documents were created by first drawing three topics randomly, with replacement, from the above list of six. A random document from each topic was then selected, and the three documents were concatenated together to form a synthetic document containing either one, two or three different "aspects." A total of 200 documents were generated in this manner.

This synthetic collection thus simulates a set of retrieved documents to answer a query, which we then wish to analyze in order to extract the aspect structure. The data was used to train aspect models using both EP and VB, fixing the number of aspects at six; 75% of the data was used for training, and the remaining 25% was used as test data.

| Aspect 1 | Aspect 2 | Aspect 3 | Aspect 4 | Aspect 5 | Aspect 6 | Aspect 1 | Aspect 2 | Aspect 3 | Aspect 4 | Aspect 5 | Aspect 6 |
|---|---|---|---|---|---|---|---|---|---|---|---|
| SAID | SAID | SAID | SAID | SAID | SAID | AGRICULTURE | REPORT | RIOT | FORMER | STORM | MANDELA |
| FOR | FOR | FOR | FOR | WAS | HE | PRICES | COULD | SEOUL | TELEDYNE | SNOW | AFRICAN |
| THAT | WAS | POLICE | HE | AT | FOR | FARMERS | MANY | MAY | CHARGES | WEATHER | ANC |
| BY | THAT | WITH | THAT | WERE | THAT | PRODUCTION | BILLION | COMMUNIST | DEFENSE | RAIN | DE |
| ON | BY | ON | WAS | FOR | WAS | GRAIN | THEM | KOREA | INDICTMENT | TEXAS | KLERK |
| WHEAT | ON | BY | ON | ON | IS | BILLION | OFFICIAL | PROTESTERS | INVESTIGATION | INCHES | ANTIAPARTHEID |
| WAS | WERE | THAT | WITH | BY | WITH | PROGRAM | LAW | OPPOSITION | OFFICIAL | WINDS | CONGRESS |
| MILLION | WITH | WAS | BY | WITH | SOUTH | SUBSIDIES | CORRUPTION | PROTESTS | DRUGS | POWER | POLITICAL |
| IS | AT | WERE | IS | THAT | BY | REPORT | DEPARTMENT | ROH | FEDERAL | SERVICE | LEADER |
| FROM | HE | AN | WERE | FROM | ON | BUSHELS | COLOMBIA | STUDENT | GUILTY | MPH | WHITE |
| AT | FROM | STUDENTS | FROM | IT | HAS | CHINA | CHARGES | ARRESTED | PROSECUTORS | DAMAGE | BLACKS |

Figure 2: The top words, sorted in order of decreasing probability, for each aspect without filtering out common words (left) and after removing them from the lists (right). As a model fit using maximum likelihood, the aspect model assigns significant mass to the common, "content-free" words. The filter lists demonstrate that the model has captured the true underlying aspects.

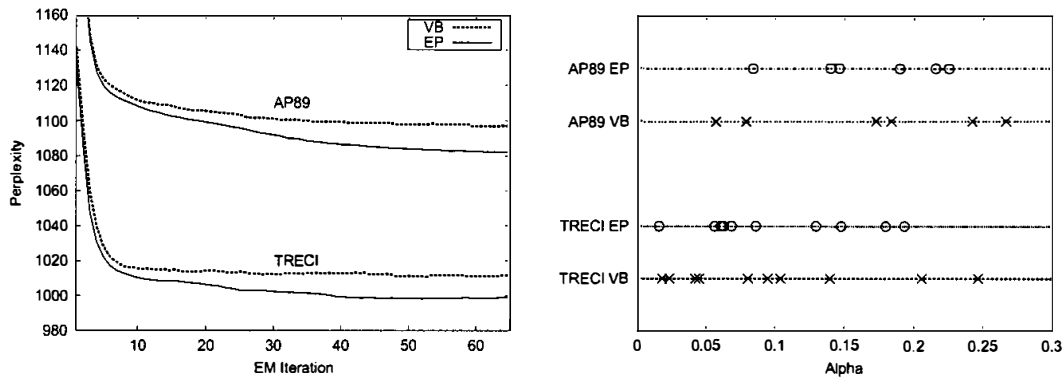

Figure 3: The left plot shows the test set perplexities as a function of EM iteration. The perplexities for the EP-trained models are lower than those of the VB-trained models. The right plot shows Dirichlet parameters, where the parameters are normalized to sum to one. The spread of the $\alpha_a$s for the VB-trained model is greater than for the EP-trained model, indicating that some of the aspects are more general (high $\alpha_a$) or specialized (low $\alpha_a$).

Figure 2 shows the top words for each aspect for the EP-trained model. Because likelihood is used as the objective function, the common, "content-free" words take up a significant portion of the probability mass—a fact that is often not acknowledged in descriptions of aspect models. As seen in this figure, the aspects model variations across documents in the distribution of common words such as SAID, FOR, and WAS. After filtering out the common words from the list, by not displaying words having a unigram probability larger than a threshold of 0.001, the most probable words that remain clearly indicate that the true underlying aspects have been captured, though some more cleanly than others. For example, aspect 1 corresponds to topic 142, and aspect 5 corresponds to topic 59.

The models are compared quantitatively using test set perplexity, $\exp(-(\sum_i \log p(d_i))/ \sum_i |d_i|)$; lower perplexity is better. The probability function (3) cannot be computed analytically, and we do not want to favor either of the two approximations, so we use importance sampling to compute perplexity. In particular, we sample $\lambda$ from the approximate posterior $\mathcal{D}(\lambda \mid \gamma)$ obtained from EP.

Figure 3 shows the test set perplexities for VB and EP; the perplexity for the EP-trained model is consistently lower than the perplexity of the VB-trained model. Based on the results of Section 6.1, we anticipate that for VB the aspects will be more extreme and specialized. This would make the Dirichlet weights $\alpha_a$ smaller for the specialized aspects, which are used infrequently, and larger for the aspects that are used in different topics or that are devoted to the common words. Plots of the Dirichlet parameters (Figure 3, center and right) show that VB results in $\alpha_a$s that are indeed more spread out towards these extremes, compared with those obtained using EP.

### 6.3 TREC Interactive Data

To compare VB and EP on real data having a mixture of aspects, this section considers documents from the TREC interactive collection (Over, 2001). The data used for this track is interesting for studying aspect models because the relevant documents have been hand labeled according to the specific aspects of a topic that they cover. Here we simply evaluate perplexities of the models.

We extracted all of the relevant documents for each of the six topics that the collection has relevance judgements for, resulting in a set of 772 documents. The average document length is 594 tokens, and the total vocabulary size is 26,319 words. As above, 75% of the data was used for training, and the remaining 25% was used for evaluating perplexities. In these experiments the speed of VB and EP are comparable.

Figure 3 shows the test set perplexity and Dirichlet parameters $\alpha_a$ for both EP and VB, trained using $A = 10$ aspects. As for the controlled TREC data, EP achieves a lower perplexity, and has aspects that are more balanced compared to those obtained using VB. We suspect that the perplexity difference on both the TREC interactive and controlled TREC data is small because the true aspects have little overlap, and thus the posterior of the mixing weights is sharply peaked.

## 7 Conclusions

The generative aspect model provides an attractive approach to modeling the variation of word probabilities across documents, making the model well suited to information retrieval and other text processing applications. This paper studied the problem of approximation methods for learning and inference in the generative aspect model, and proposed an algorithm based on Expectation-Propagation as an alternative to the variational method adopted by Blei et al. (2001). Experiments on synthetic data showed that simple variational inference can lead to inaccurate inferences and biased learning, while Expectation-Propagation can lead to more accurate inferences. Experiments on TREC data show that Expectation-Propagation achieves lower test set perplexity. We attribute this to the fact that the Jensen bound used by the variational method is inadequate for representing how 'peaky' versus 'spread out' is the posterior on $\lambda$, which happens to be crucial for good parameter estimates. Because there is a separate $\lambda$ for each document, this deficiency is not minimized by additional documents, but rather compounded.

## Acknowledgements

We thank Cheng Zhai and Zoubin Ghahramani for assistance and helpful discussions. Portions of this work arose from the first author's internship with Andrew McCallum at JustResearch.

## Appendix: Updating $p(w \mid a)$

The update for $p(w \mid a)$ requires approximating the integral

$$\int_\Delta q_i(\lambda) \frac{p(w \mid a)\lambda_a}{\sum_b p(w \mid b)\lambda_b} d\lambda$$

$$= \int_\Delta \mathcal{D}(\lambda \mid \gamma) \frac{p(w \mid a)\lambda_a}{\sum_b p(w \mid b)\lambda_b} \quad (36)$$

$$= \frac{p(w \mid a)\gamma_a}{\sum_b \gamma_b} \int_\Delta \mathcal{D}(\lambda \mid \tilde{\gamma}) \frac{1}{\sum_b p(w \mid b)\lambda_b} \quad (37)$$

where

$$\tilde{\gamma}_b = \begin{cases} \gamma_b + 1 & \text{if } b = a \\ \gamma_b & \text{otherwise} \end{cases} \quad (38)$$

This reduces to an expectation under a Dirichlet density. Any expectation $E[f(\lambda)]$ can be approximated via a Taylor-expansion of $f$ about $E[\lambda]$, as follows:

$$f(\lambda) \approx f(E[\lambda]) + f'(E[\lambda])^T(\lambda - E[\lambda])$$
$$+ \frac{1}{2}(\lambda - E[\lambda])^T f''(E[\lambda])(\lambda - E[\lambda]) \quad (39)$$

$$E[f(\lambda)] \approx f(E[\lambda]) + \frac{1}{2}\text{tr}(f''(E[\lambda])\text{Var}(\lambda)) \quad (40)$$

where $\text{Var}(\lambda)$ is the covariance matrix of $\lambda$. In our case,

$$f(\lambda) = \frac{1}{\sum_b p(w \mid b)\lambda_b} \quad (41)$$

$$E[\lambda_b] = \frac{\tilde{\gamma}_b}{\sum_s \tilde{\gamma}_s} = m_{iab} \quad (42)$$

and after some algebra we reach (34). A second-order approximation works well for $f$ because it curves only slightly for realistic values of $\lambda$.